\documentclass[letterpaper]{article} 
\usepackage
{aaai2027}  
\usepackage[hyphens]{url}  
\usepackage{graphicx} 
\usepackage{natbib}  
\usepackage{caption} 
\usepackage{booktabs}

\usepackage{amsmath}
\usepackage{amssymb}
\usepackage{multirow}
\usepackage{tikz}
\usepackage{float}

\usepackage{xr}
\nocopyright

\title{Transformers Struggle to Use Their Emergent World Models: Revisiting the Tower of Hanoi, and the Illusion of Thinking}

\author{
Devin Pereira\textsuperscript{\rm 1,}\textsuperscript{\rm 2},
Willem Zuidema\textsuperscript{\rm 3,}\textsuperscript{\rm 2}\corresponding
}
\affiliations{
\textsuperscript{\rm 1}Artificial Intelligence Program, University of Amsterdam\\
\textsuperscript{\rm 2}
ELLIS Unit Amsterdam\\
\textsuperscript{\rm 3}
Institute for Logic Language and Computation, 
University of Amsterdam \\Science Park 900, 1098XH Amsterdam, the Netherlands \\
zuidema@uva.nl

}

\begin{document}

\maketitle

\begin{abstract}
The Tower of Hanoi is a simple planning puzzle that in prior work has proven challenging for large reasoning models (LRMs). Current models solve the standard formulation of the puzzle, but still struggle with the flat-to-flat variant (where initial and goal states are not restricted to have all rings on a single peg). This paper presents an in-depth study of how both small, in-house Transformers and large, third-party LRMs solve this task. To understand the failures mechanistically, we first train small Transformers from scratch on precomputed solution traces. Using a variety of interpretability techniques, we show that these Transformers develop an emergent world model: a linearly decodable, geometrically faithful representation of the puzzle's state space (the Sierpi\'nski triangle), that is causally involved in solving the puzzles. Second, we return to the large LLMs and apply our techniques to two frontier reasoning models, Qwen3.6-27B and DeepSeek-R1-Distill-Qwen-32B, that attempt to solve the task through extended chain-of-thought. Surprisingly, we find that both models encode the Sierpi\'nski world model near-perfectly at the end of the prompt, and yet fail at the majority of tasks when $N\geq4$. We locate the source of this failure in the decaying representation of the world model. We probe for the representation at different stages during planning, and establish causality by showing that performance can be improved by injecting the prompt-time representation at inference. The failure of the models is thus one of maintenance of the required representations, not their absence, and performance is at least partially recoverable. These results thus reframe the reported collapse in performance from prior work: current Large Reasoning Models build a world model, and then lose it. 
\end{abstract}

\section{Introduction}
\label{sec:introduction}

Current large reasoning models are capable of drafting mathematical proofs, writing complex code, and scoring at the frontier of graduate examinations. And yet, they cannot---at least without tool use---reliably solve a puzzle popularly used to assess children's planning abilities. \citet{shojaee2025illusion} studied the abilities of such models on the Tower of Hanoi task, where a stack of rings of different sizes needs to be moved from the source peg to the goal peg, subject to the constraints that only the topmost ring on a peg may be moved and that a larger ring may never be placed on top of a smaller one. \citeauthor{shojaee2025illusion} report that the accuracy of these models falls off a cliff once the puzzle grows past a handful of disks, and, stranger still, their reasoning traces get shorter at exactly the point where the problem gets harder. The paper named this phenomenon the ``illusion of thinking'', kicking off an intense debate about the generalizability of the progress on reasoning~\citep{varela_rethinking_2025,meyerson_solving_2025,zhao_is_2026}. Despite many follow-ups, the phenomenon has not been explained mechanistically: we know that these models fail, but we do not know what inside the model is failing.

In this paper\footnote{This paper is based on the unpublished master thesis of Devin Pereira (2026), University of Amsterdam.
}, we present an in-depth case study into the mechanisms by which both small Transformers and large LRMs (try to) solve the Tower of Hanoi task. We start by describing two variants of the task and our baseline results. We then, in Section~\ref{sec:toy}, study our in-house trained Transformers, asking whether they show evidence of having acquired an ``emergent world model''~\citep{nanda2023emergent}. In Section~\ref{sec:lrm} we return to LRMs and show evidence that they build up the same emergent world models for this task, but lose them when generating the actual plan.

\section{The Tower of Hanoi Puzzle}
\label{sec:toh}

The Tower of Hanoi is a simple planning puzzle, invented by French mathematician Fran\c{c}ois \'Edouard Anatole Lucas in 1883, and introduced into AI and cognitive science by Herbert Simon~\citep{simon_functional_1975}. In the standard set-up, it involves $M=3$ pegs, and $N=4$ rings of different sizes. The player can move one ring at a time from one peg to another, subject to the constraints that only the top ring on a stack can be moved and that a larger ring cannot be placed on top of a smaller ring. The task of the player is to find a correct (and possibly shortest) sequence of legal moves from an initial configuration to a goal configuration.

\begin{figure}[t]
  \centering
\begin{tabular}{lcl}
\begin{tikzpicture}[scale=0.2]
  \draw[fill=brown!60!black] (1.5,-0.2) rectangle (10.5,0.2);  
  \foreach \x in {2, 6, 10} {\draw[fill=gray!50!black] (\x-0.1,0) rectangle (\x+0.1,4);} 
  \node at (2,-1.0) {1}; \node at (6,-1.0) {2}; \node at (10,-1.0) {3}; 
  \draw[fill=red!70!black, rounded corners=2pt] (1.5,2.6) rectangle (2.5,3.1);   
  \draw[fill=blue!70!black, rounded corners=2pt] (1.0,1.9) rectangle (3.0,2.4);  
  \draw[fill=green!70!black, rounded corners=2pt] (0.5,1.2) rectangle (3.5,1.7);
  \draw[fill=black, rounded corners=2pt] (0,0.5) rectangle (4,1.0);
\end{tikzpicture} & $\longrightarrow$ &
\begin{tikzpicture}[scale=0.2]
  \draw[fill=brown!60!black] (1.5,-0.2) rectangle (10.5,0.2);  
  \foreach \x in {2, 6, 10} {\draw[fill=gray!50!black] (\x-0.1,0) rectangle (\x+0.1,4);} 
  \node at (2,-1.0) {1}; \node at (6,-1.0) {2}; \node at (10,-1.0) {3}; 
  \draw[fill=red!70!black, rounded corners=2pt] (9.5,2.6) rectangle (10.5,3.1);   
  \draw[fill=blue!70!black, rounded corners=2pt] (9.0,1.9) rectangle (11.0,2.4);  
  \draw[fill=green!70!black, rounded corners=2pt] (8.5,1.2) rectangle (11.5,1.7);
  \draw[fill=black, rounded corners=2pt] (8.0,0.5) rectangle (12,1.0);
\end{tikzpicture} 
\\
\multicolumn{3}{c}{``tower-to-tower''}
\\\\
\begin{tikzpicture}[scale=0.2]
  \draw[fill=brown!60!black] (1.5,-0.2) rectangle (10.5,0.2);  
  \foreach \x in {2, 6, 10} {\draw[fill=gray!50!black] (\x-0.1,0) rectangle (\x+0.1,4);} 
  \node at (2,-1.0) {1}; \node at (6,-1.0) {2}; \node at (10,-1.0) {3}; 
  \draw[fill=red!70!black, rounded corners=2pt] (1.5,1.2) rectangle (2.5,1.7);   
  \draw[fill=blue!70!black, rounded corners=2pt] (1.0,0.5) rectangle (3.0,1.0);  
  \draw[fill=green!70!black, rounded corners=2pt] (8.5,0.5) rectangle (11.5,1.0);
  \draw[fill=black, rounded corners=2pt] (4,0.5) rectangle (8,1.0);
\end{tikzpicture} & $\longrightarrow$ &
 \begin{tikzpicture}[scale=0.2]
  \draw[fill=brown!60!black] (1.5,-0.2) rectangle (10.5,0.2);  
  \foreach \x in {2, 6, 10} {\draw[fill=gray!50!black] (\x-0.1,0) rectangle (\x+0.1,4);} 
  \node at (2,-1.0) {1}; \node at (6,-1.0) {2}; \node at (10,-1.0) {3}; 
  \draw[fill=red!70!black, rounded corners=2pt] (1.5,0.5) rectangle (2.5,1.0);   
  \draw[fill=blue!70!black, rounded corners=2pt] (9.0,1.9) rectangle (11.0,2.4);  
  \draw[fill=green!70!black, rounded corners=2pt] (8.5,1.2) rectangle (11.5,1.7);
  \draw[fill=black, rounded corners=2pt] (8.0,0.5) rectangle (12,1.0);
\end{tikzpicture}\\
\multicolumn{3}{c}{``flat-to-flat''}
\end{tabular}
    \caption{Tower-to-tower (top) and flat-to-flat (bottom) variants of the 4-disk Tower of Hanoi. In the flat-to-flat case both the initial and goal configurations have disks spread across pegs, so the optimal path depends on the pair $(s_I,s_G)$ and cannot be produced by a memorised recursive template.}
    \label{fig:toh-variants}
\end{figure}
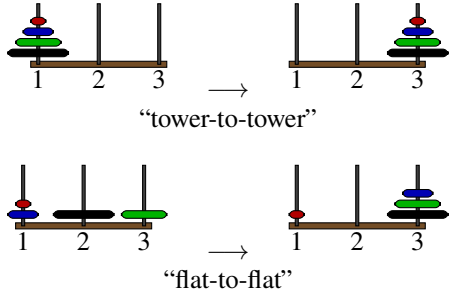

The puzzle---in the most well-known version where both initial and goal configurations have all rings on a single peg (``tower-to-tower'')---is a popular programming assignment, as it allows for an elegant recursive implementation where the solution to the problem of moving a tower of $N$ rings (to the goal peg) includes the solution of moving a tower of $N-1$ rings (to the non-goal peg). The puzzle---in a variant where initial and goal configurations can also have rings on multiple pegs (``tower-to-flat'', ``flat-to-tower'', ``flat-to-flat'')---is also a popular psychological test, where performance of adults and children of various ages is thought to correlate with executive function~\citep{welsh_development_2001}. Figure~\ref{fig:toh-variants} shows the two main variants for $N=4$ disks. In the classical tower-to-tower variant all rings start and end on a single peg; in the flat-to-flat variant the rings are spread across pegs in both the initial and the goal state.

\begin{figure}[H]
  \centering
  \includegraphics[width=\linewidth,trim=20 30 280 20,clip]{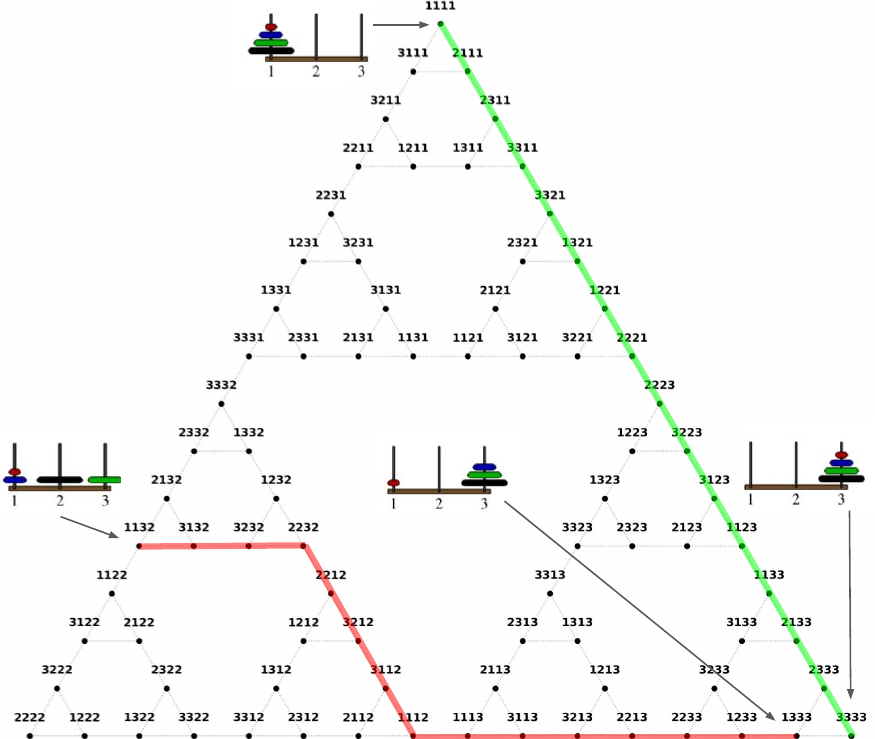}
  \caption{The state space of the 4-disk Tower of Hanoi forms a Sierpi\'nski triangle. Each node is labelled by a triple indicating the peg on which each disk rests, ordered from smallest to largest. The three sub-triangles partition the space according to the position of the largest disk.}
  \label{fig:toh-statespace}
\end{figure}

Finally, in (recreational) mathematics the puzzle is known for its state space: all possible configurations can be arranged in a shape known as the Sierpi\'nski triangle (Sierpi\'nski gasket; Figure~\ref{fig:toh-statespace}). This triangle is arranged such that all legal moves correspond to transitions between direct neighbours, and the points of the triangle correspond to configurations with all rings on one peg.

In current AI, the Tower of Hanoi puzzle played a major role in the ``Illusion of Thinking'' paper~\citep{shojaee2025illusion} and the ensuing debate. That paper studied the performance of some of the 2025 frontier reasoning models on four tasks, including the tower-to-tower one. The authors report that, despite prompt optimization, performance drops dramatically for $N\geq4$ rings. In contrast, we evaluated a range of 2026 frontier models (without tool use, using the prompt from \citealp[p17,18]{shojaee2025illusion}), and find that all of them solve standard tower-to-tower tasks near-perfectly across $n\in\{3,4,5\}$ (DeepSeek-R1 24/25, Kimi-K2-Think 23/25, gpt-oss-120b 25/25, Qwen3.6-27B 24/25). 
In our experiments, only the distilled DeepSeek-R1-Distill-Qwen-32B failed on most runs (solving just 1/25).

While the standard tower-to-tower task is thus saturated for current reasoning models, the flat-to-flat formulation is far harder (Table~\ref{tab:flatflat}). Even the strongest models solve only about half of instances optimally, and accuracy falls steeply with disk count; extending the sweep for Qwen3.6-27B, the optimal count drops to $4/33$ at $n{=}6$ and $2/33$ at $n{=}7$. Notably Qwen3.6-27B generates longer traces than DeepSeek-R1-Distill-Qwen-32B on average ($\approx19{,}000$ vs $\approx14{,}000$ tokens), a fact we return to for the steering asymmetry.

\begin{table}[t]
  \centering
  \setlength{\tabcolsep}{3pt}
  \caption{Baseline flat-to-flat accuracy: instances solved optimally per model, out of the number sampled at each disk size (column headers; 100 total for $n{=}3$--$5$), under a 32k output-token budget (which caps the answer only, with provider-side reasoning beyond it). $n{=}6,7$ were evaluated only for Qwen3.6-27B (33 each); ``--'' marks sizes not run for a model.}
  \label{tab:flatflat}
  \small
  \begin{tabular}{lcccccc}
    \toprule
    \textbf{Model} & \shortstack{\textbf{3}\\{\scriptsize/34}} & \shortstack{\textbf{4}\\{\scriptsize/33}} & \shortstack{\textbf{5}\\{\scriptsize/33}} & \shortstack{\textbf{6}\\{\scriptsize/33}} & \shortstack{\textbf{7}\\{\scriptsize/33}} & \shortstack{\textbf{Total}\\{\scriptsize(3--5)}} \\
    \midrule
    DeepSeek-R1                  & 24 & 12 & 4  & -- & -- & \textbf{40} \\
    Kimi-K2-Think                & 19 & 3  & 3  & -- & -- & \textbf{25} \\
    gpt-oss-120b                 & 26 & 12 & 13 & -- & -- & \textbf{51} \\
    \midrule
    DeepSeek-R1-Distill  & 6  & 0  & 1  & -- & -- & \textbf{7} \\
    Qwen3.6-27B                  & 28 & 14 & 9  & 4 & 2 & \textbf{51} \\
    \bottomrule
  \end{tabular}
\end{table}

\section{Related Work}
\label{sec:related-work}

\subsubsection{Emergent world models} 
Transformer sequence models can develop internal representations of the latent state they predict over, even when trained only on next-token prediction~\citep{li2022emergent, toshniwal2022chess}. For instance, \citet{li2022emergent} show they can decode board positions from the internal states of a model trained on just sequences of moves in the game of Othello. Such 'emergent world models' have been demonstrated exclusively in small Transformers trained on games~\citep{li2022emergent, toshniwal2022chess} or path prediction in mazes~\citep{spies2025transformers}. 

\paragraph{Reasoning models} In the last few years, much attention is focused on large reasoning models, that are trained to generate an extensive chain of thought before generating the final answer \citep[e.g.,][]{guo2025deepseek}. The chain of thought is often taken as the explanation for the models behavior, but is not guaranteed to be faithful to the true underlying mechanisms \cite{anthropic_reasoning_2025,arcuschin_chain--thought_2026}. Analysis of how frontier (agentic) reasoning model arrive at their output is made difficult by the size of these models, as well as by the use of external tools at various steps in their operation \cite{ke_survey_2025,anthropic_how_2025,openai_harness_2026}. To which components of these models their evident reasoning successes must be attributed remains a topic of debate \citep[e.g.,][]{valmeekam2023planning}. 
Whether large reasoning models show evidence of emergent world models is yet to be established: there is a paucity of research on whether a reasoning model's internal representation of planning problems is what determines success.

\subsubsection{Probing and intervening on representations.}
We will test for encoded world models with linear probes~\citep{belinkov2022probing}, motivated by the linear representation hypothesis~\citep{mikolov2013linguistic, park2023linear}, and remain mindful that some features are nonlinear or genuinely multi-dimensional~\citep{csordas2024recurrent, engels2025not}. 
Decodability does not confirm that models actually use this representation; approaches to demonstrate causal involvement include activation patching~\citep{vig2020investigating, meng2022locating, wang2022interpretability} and  activation steering~\citep{turner2023steering, subramani2022extracting, hernandez2023inspecting}. The closest methodological precedent to the current paper is presented in \citep{li2022emergent, spies2025transformers}, where sequence-only training is combined with probing and causal edits. 

\section{Small Transformers}
\label{sec:toy}

We train a GPT-2-style decoder-only Transformer~\citep{radford2019language} on tokenised flat-to-flat solution traces (generated symbolically) with a cross-entropy loss masked to the move tokens. All experiments use the $n=4$ Tower of Hanoi, whose 81 valid configurations yield $6{,}480$ ordered $(s_I,s_G)$ pairs with $s_I\neq s_G$, split 80/20 into $5{,}184$ training and $1{,}296$ validation problems. The model has 6 layers, hidden dimension 128, 4 heads, and is trained for 50 epochs. It reaches 99.2\% token-level and 93.2\% sequence-level validation accuracy. 
Each problem is serialised as the start configuration, the goal configuration, and the move sequence, with the cross-entropy loss masked to the move suffix:

{\footnotesize
\begin{verbatim}
tokens: BOS P0 P0 P0 P0 SEP P0 P1 P0 P0 SEP
ids:     9   0  0  0  0  10  0  1  0  0  10
        |_____ prefix (conditioning) _____|

tokens: M02 M01 M20 ... EOS
ids:      4   3   7      11
        |   loss target   |
\end{verbatim}
}

\subsection{The Joint State Is Linearly Decodable at SEP}

\begin{figure*}[t]
  \centering
  \includegraphics[width=0.24\linewidth]{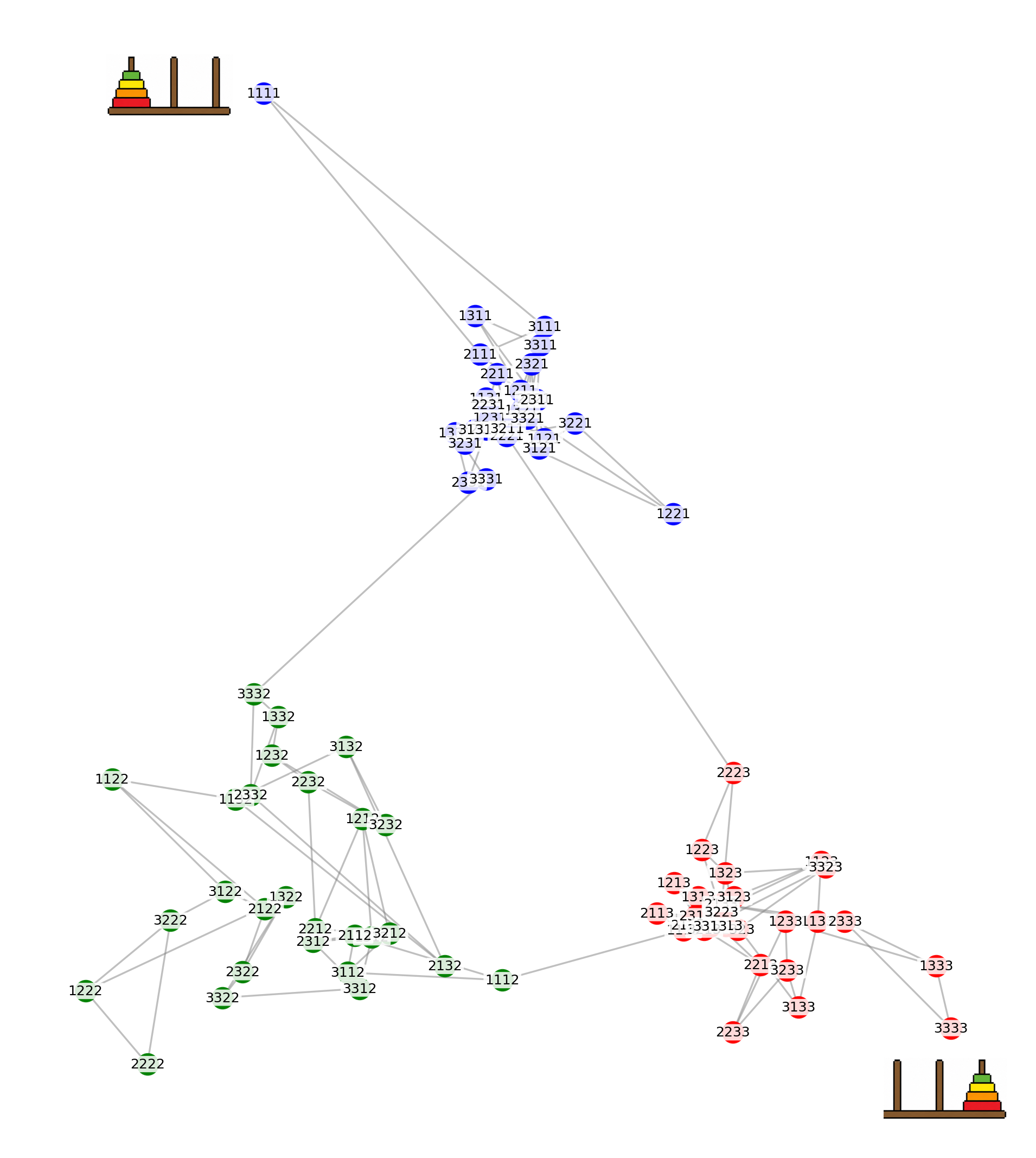}\hfill
  \includegraphics[width=0.24\linewidth]{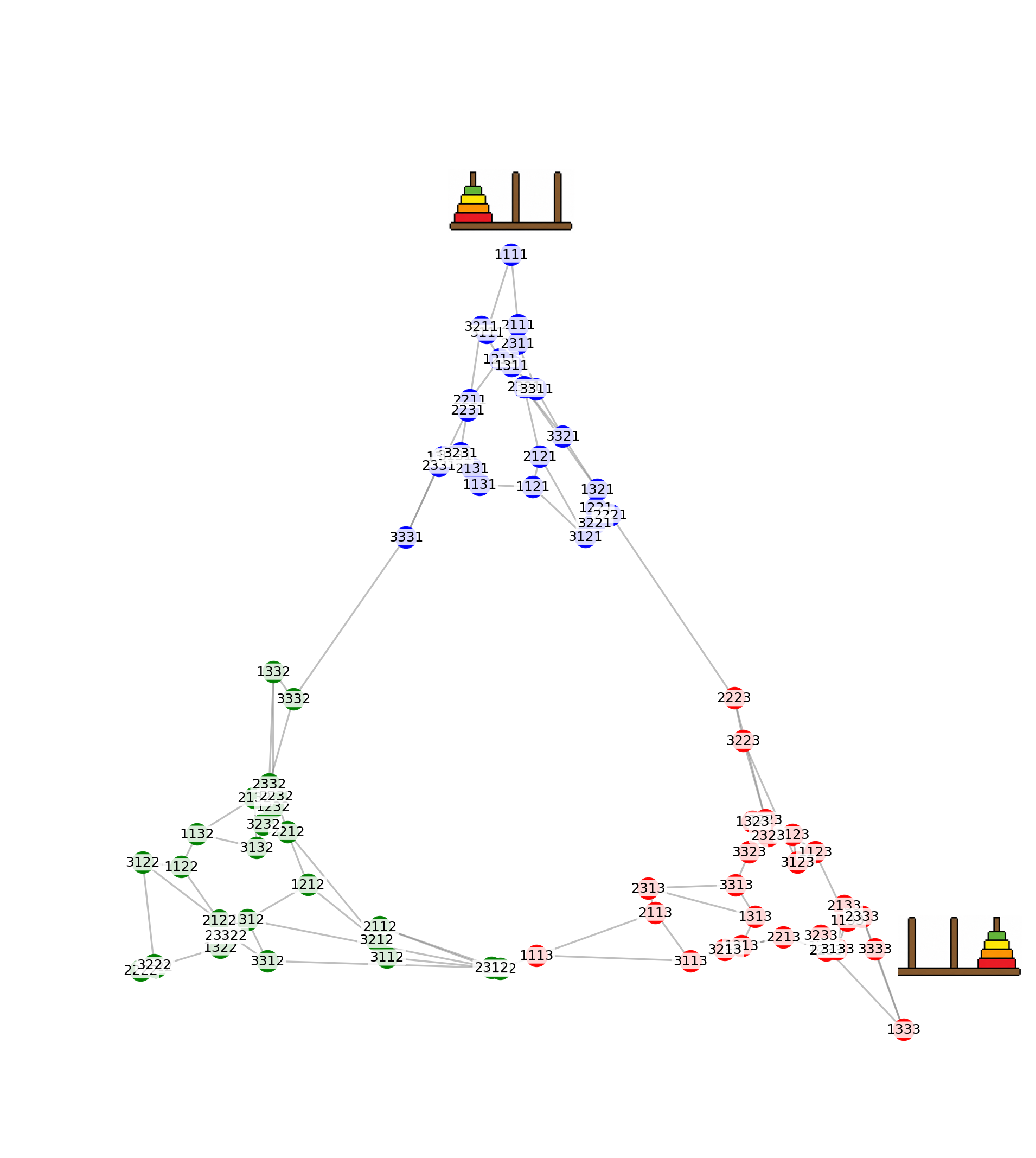}\hfill
  \includegraphics[width=0.24\linewidth]{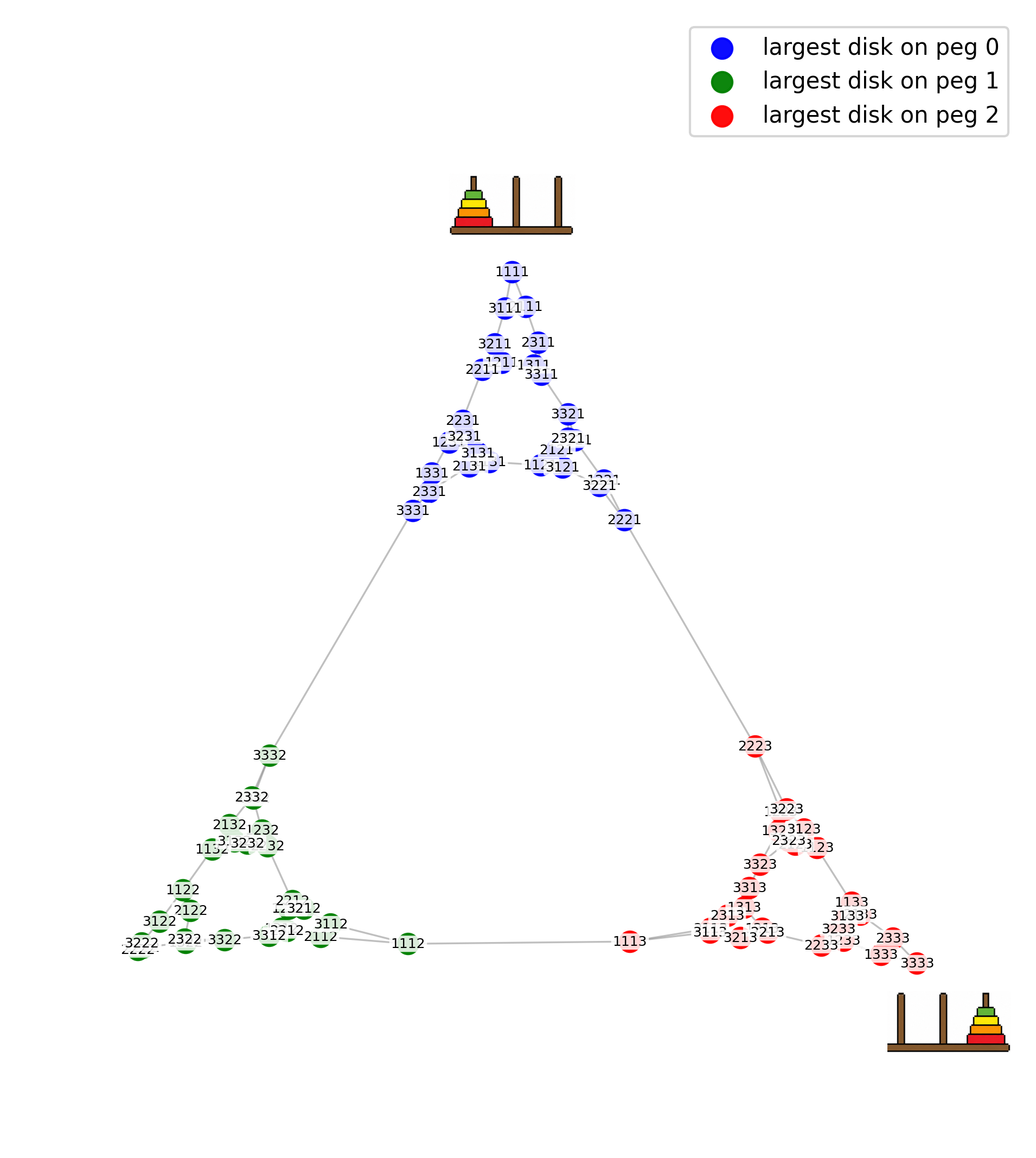}\hfill
  \includegraphics[width=0.24\linewidth,trim=45 25 10 10,clip]{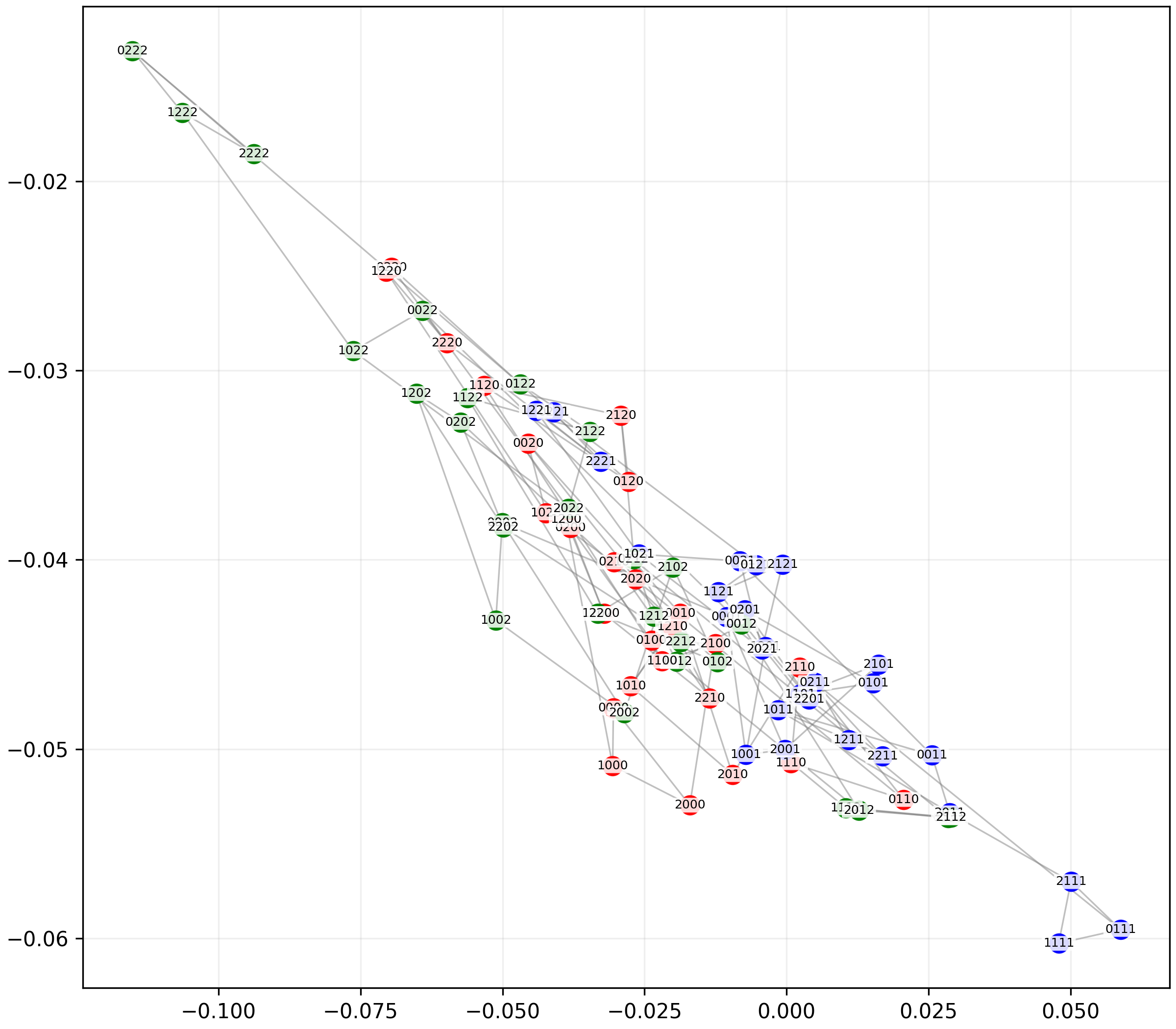}
  \caption{2D output of the distance-matching probe on the raw residual stream of the SEP token at layer 5, over training (from left to right: epochs 5, 25, 50), and at a MOVE token (rightmost). Each point is one of the 81 four-disk configurations, coloured by the largest disk's position; grey edges connect graph-adjacent configurations. At the SEP token, the three largest-disk sub-triangles separate into the Sierpi\'nski geometry, with cross-cluster gaps wider than the within-cluster edges.}
  \label{fig:probe-2d}
\end{figure*}

To test whether the Sierpi\'nski state space is (linearly) encoded, we train a linear probe $f_\phi:\mathbb{R}^d\!\to\!\mathbb{R}^2$ that maps a hidden state to a 2D embedding whose pairwise Euclidean distances match the graph distances $d_G$ between configurations:
\begin{equation}\label{eq:distance-probe}
  \mathcal{L}_{\text{probe}}(\phi) = \frac{1}{|\mathcal{S}|^2} \sum_{s, s' \in \mathcal{S}} \bigl( \|f_\phi(h_s) - f_\phi(h_{s'})\|_2 - d_G(s, s') \bigr)^2,
\end{equation}
where $\mathcal{S}$ is the set of all 81 four-disk configurations. The state of all four disks is read out at the separator token (SEP) that sits between the problem and the solution. Probes are trained at each of the 6 layers. We find that over training the Sierpi\'nski structure becomes more and more apparent (Figure~\ref{fig:probe-2d}).

Probe results are also reported quantitatively by the Spearman ($\rho$) and Pearson ($r$) correlations between predicted and true pairwise distances, and by nearest-state retrieval accuracy (the fraction of configurations whose nearest embedding neighbour is the true nearest graph neighbour). A high $\rho$ with lower $r$ signals an order-preserving but non-uniformly scaled embedding. The probe applied to the residual stream at the SEP token recovers the joint four-disk configuration with high fidelity (Table~\ref{tab:sep-probe}): Spearman correlation reaches $0.94$ at the best layer (layer 5) and Pearson $0.90$, with the geometry assembled by middle depth. The persistent $\rho{-}r$ gap of $3$--$4$ points indicates an embedding that preserves the order of graph distances but inflates the separation between the three sub-triangles defined by the largest disk's position. 

We also train probes on the residual streams at the move tokens. These probes use the $49{,}450$ move-token activations from optimally solved problems, but, by contrast, fail to recover the joint state (best $\rho=0.78$ with a 15-token window)
, even though the same state is perfectly decodable 
a few positions earlier, motivating a per-disk analysis. 

\begin{table}[t]
  \centering
  \caption{Linear probes reveal a near-perfect representation of the state space at the SEP token's residual stream. \emph{Left:} SEP-token distance-matching probe at epoch 50, evaluated with Spearman \textbf{$\rho$} and Pearson correlation \textbf{$r$} on all 81 configurations, by layer, peaking at layer 5. Right: per-disk classification accuracy (best layer, 5); perfect at SEP, larger disks degrade at move tokens.}
  \label{tab:sep-probe}
  \resizebox{\linewidth}{!}{
\centerline{
  \begin{tabular}{rrr}
    \toprule
    \textbf{Layer} & \textbf{$\rho$} & \textbf{$r$} \\ 
    \midrule
    6 & {0.934} & 0.898 \\
    5 & \textbf{0.938} & \textbf{0.902} \\
    4 & {0.936} & 0.899 \\
    3 & {0.931} & 0.895 \\
    2 & {0.921} & 0.878 \\
    1 & {0.829} & 0.814 \\
    \bottomrule
  \end{tabular}
  \hspace{0.3cm}
  \label{tab:per-disk}
  \begin{tabular}{lll}
    \toprule
    \textbf{Position} & SEP & Move \\ \midrule
    {Disk 0} & {100.0\%} & 99.98\% \\
    {Disk 1} & {100.0\%} & 99.96\% \\
    {Disk 2} & {100.0\%} & 90.70\% \\
    {Disk 3} & {100.0\%} & 79.25\% \\ \hline \\
    {Mean} & {100.0\%} & 92.47\% \\
    \bottomrule
  \end{tabular}
  }
  }
\end{table}

\subsection{A Unified-to-Factored Shift Between Planning and Execution}

To characterise how the state is organised, we train one logistic probe per disk $k$ (weight matrix $W_k\in\mathbb{R}^{3\times d}$) predicting its peg, and treat the row span of $W_k$ as that disk's encoding subspace. The \emph{principal angle} between two such subspaces (the smallest angle between any pair of unit vectors drawn from them, via the SVD of their orthonormal bases) measures representational format. Angles near $0^\circ$ indicate a \emph{unified} encoding in a shared subspace, angles near $90^\circ$ a \emph{factored}, mutually orthogonal one.

The per-disk probes show the state is present at both positions but organised differently (Table~\ref{tab:per-disk}). At SEP every disk is recovered perfectly; at move tokens the two small disks remain near-perfect while the larger disks degrade (to $90.7\%$ and $79.3\%$), as the model loses track of infrequently moved disks. Principal angles make the format explicit. Between disk subspaces, they average $54.7^\circ$ at SEP (a \emph{unified}, overlapping encoding that supports the joint Sierpi\'nski geometry) but rise to $76.5^\circ$ at move tokens (a \emph{factored}, near-orthogonal encoding from which each disk can be read independently). The joint distance structure is thus not the sum of the per-disk parts: it lives in the overlapping geometry that the distance probe captures and the per-disk probes discard.

\subsection{The Representation Is Causally Used}

To test causal involvement of the representations in the residual stream at the SEP token, we patch~\citep{meng2022locating} the SEP activation: running the model on a corrupted configuration $s_\text{corrupt}$ but substituting the SEP activation from a clean pass on $s_\text{clean}$, and classifying the emitted sequence as \textsc{full transfer} (matches $s_\text{clean}$), \textsc{partial} (some disks follow the donor), \textsc{unchanged} (matches $s_\text{corrupt}$), \textsc{novel} (optimal for a third configuration), or \textsc{disrupted} (solves neither).

Substituting a donor's SEP activation transfers state to the recipient's output in a large majority of pairs (Table~\ref{tab:patching}). \textsc{Partial} transfer dominates at every layer ($64$--$79\%$), where $K\approx1.45$ of four disks on average are copied from the donor's solution. \textsc{Full} transfer is present ($\sim6\%$), and \textsc{novel} outcomes are rare ($\leq0.4\%$), so outputs stay tied to donor or recipient. The \textsc{disrupted} rate falls with depth, from $12.4\%$ at layer 4 to zero at layer 6, i.e.\ late-layer SEP activations admit clean replacement. We conclude that the influence is causal but graded. 

Note that in these experiments we intervene on the whole residual stream at each layer, not surgically on the 2D emergent-world-model representation; we leave such more surgical experiments for future work, but take the evidence of a causal role of the layer 5 and 6 residual streams at SEP as sufficient inspiration to return to large reasoning models.

\begin{table}[t]
  \centering
  \setlength{\tabcolsep}{3.5pt}
  \caption{Activation patching at the SEP token across all donor--recipient validation pairs. \textsc{Partial} transfer dominates; \textsc{disrupted} falls to zero with depth.}
  \label{tab:patching}
  \begin{tabular}{rrrrrr}
    \toprule
    \textbf{Layer} & \textbf{Full} & \textbf{Partial ($K$)} & \textbf{Unch.} & \textbf{Novel} & \textbf{Disr.} \\
    \midrule
    4 & 5.75\% & \textbf{63.89\% (1.46)} & 17.56\% & 0.37\% & 12.42\% \\
    5 & 6.00\% & \textbf{71.77\% (1.45)} & 16.49\% & 0.11\% &  5.63\% \\
    6 & 5.89\% & \textbf{78.74\% (1.44)} & 15.37\% & 0.00\% &  0.00\% \\
    \bottomrule
  \end{tabular}
\end{table}

\section{Large Reasoning Models}
\label{sec:lrm}

We perform extensive experiments with two very capable, open-weight LRMs that can still be run on locally available hardware: Qwen3.6-27B and DeepSeek-R1-Distill. We use the optimized prompt from \citet{shojaee2025illusion}, and run the distance-matching and per-disk probes from section~\ref{sec:toy}. We apply the prompts at three positions in a single forward pass: \textbf{Position A}, the final prompt token (the LRM analogue of SEP, before any reasoning); \textbf{Position B}, the last token before the emitted move list (the commitment point, after the full chain-of-thought); and \textbf{Position C}, the move-emission tokens (one sample per move).

We probe two open-weight reasoning models that emit explicit \texttt{<think>} chains: \mbox{Qwen3.6-27B} and \mbox{DeepSeek-R1-Distill-Qwen-32B} (both 64 layers, hidden dimension $5{,}120$). Probing is applied at 17 layer indices (layer 1 and every fourth from 4 to 64). Problem instances are the 81 reachable configurations paired with a fixed goal ($N=81$ for Position A). Solutions are classified \textsc{optimal}, \textsc{suboptimal} (legal, goal-reaching, over-long), \textsc{incorrect} (legal, goal-missing), or \textsc{illegal} (a rule violation or unparseable output).  
For baselines we follow the protocol of \citet{shojaee2025illusion}, varying only the generation token budget. 

\subsubsection{Position A: a world model at the end of the prompt.}
At Position A the distance-matching probe recovers the configuration to essentially perfect accuracy in both models: from middle depth onward Spearman reaches $0.935$ and nearest-state accuracy $1.00$, identical to the toy SEP token, and the same $\rho{-}r$ asymmetry is preserved. The representation is largely absent at the first layer, but is assembled by layer 8, then plateaus; DeepSeek-R1-Distill-Qwen-32B reaches $\rho=0.936$ with perfect retrieval by middle depth as well. That a 27B pre-trained reasoning model and a 6-layer Transformer trained from scratch encode the configuration with identical fidelity is the strongest evidence that the world model is a property of the task, not of scale or training.

\subsubsection{Positions B and C: degradation through generation.}
A faithful prompt-time representation does not translate into a solution. With a maximal reasoning budget, Qwen3.6-27B solves only $51\%$ optimally and the distilled model far fewer. Tracking the probe through generation locates the gap (Table~\ref{tab:posBC}). At the commitment point (Position B) the joint configuration is not as easily decodable. For Qwen3.6-27B the global geometry is roughly preserved ($\rho\approx0.92$) but nearest-state accuracy drops and the per-disk classifiers collapse to near chance. DeepSeek-R1-Distill-Qwen-32B degrades further on both axes ($\rho$ to $0.74$--$0.81$). During move emission (Position C) the trend of a more factored, per-disk encoding partially returns. This is the large-model counterpart of the toy unified-to-factored shift: a \emph{unified} geometry where planning happens, a \emph{factored} one where execution happens. The degradation appears even on solved problems, therefore maintaining a clean configuration across a long trace is hard even when the model ultimately succeeds.

\begin{table}[t]
  \centering
  \setlength{\tabcolsep}{4pt}
  \caption{Probing Qwen3.6-27B through generation. D0--D3 are per-disk probe accuracies.}
  \label{tab:posBC}
  \begin{tabular}{llrrrrrr}
    \toprule
    \textbf{Pos} & \textbf{Layer} & \textbf{Spear.} & \textbf{Acc} & \textbf{D0} & \textbf{D1} & \textbf{D2} & \textbf{D3} \\
    \midrule
    A & 24 & 0.935 & \textbf{1.00} & 0.75 & 0.75 & 0.80 & 0.90 \\
    A & 36 & 0.935 & \textbf{1.00} & 0.59 & 0.69 & 0.75 & 0.95 \\
    A & 48 & 0.935 & \textbf{1.00} & 0.48 & 0.57 & 0.70 & 0.95 \\
    \midrule
    B & 24 & \textbf{0.915} & 0.83 & 0.33 & 0.37 & 0.42 & 0.43 \\
    B & 36 & \textbf{0.915} & 0.75 & 0.33 & 0.33 & 0.41 & 0.34 \\
    B & 48 & \textbf{0.920} & 0.82 & 0.34 & 0.36 & 0.42 & 0.43 \\
    \midrule
    C & 24 & 0.802 & 0.41 & 0.80 & 0.70 & 0.75 & \textbf{0.89} \\
    C & 36 & 0.813 & 0.64 & 0.58 & 0.62 & 0.60 & \textbf{0.84} \\
    C & 48 & \textbf{0.847} & 0.69 & 0.49 & 0.58 & 0.55 & 0.78 \\
    \bottomrule
  \end{tabular}
\end{table}

\subsubsection{Restoring the world model recovers performance.}
If world-model degradation is the bottleneck, restoring the clean Position-A representation during generation should help. To test causal use we apply activation steering~\citep{turner2023steering} at layer $\ell=28$. We cache the layer-$\ell$ Position-A activation $h_{\text{prompt}}(s)$ for each of the 81 configurations and their mean $\bar h$, and at every generated token (\emph{always-steer}) add a unit steering vector toward the clean activation of the board's current configuration $s_t$:
\begin{equation}\label{eq:steering}
  \tilde{h}^{(\ell)} = h^{(\ell)} + \alpha \cdot \frac{h_{\text{prompt}}(s_t) - \bar{h}}{\|h_{\text{prompt}}(s_t) - \bar{h}\|_2},
\end{equation}
with injection strength $\alpha$ swept. The current configuration $s_t$ is maintained by a symbolic tracker running alongside decoding: whenever a new \texttt{]} appears in the output it re-scans for the last \texttt{moves\,=\,[\,\dots\,]} block (matched by the regular expression \texttt{moves\textbackslash s*=\textbackslash s*\textbackslash[}), parses each \texttt{[disk, from, to]} triple, and replays the legal prefix from the start state; when the replayed board changes, the steering target is re-pointed at the new configuration's clean activation. In the \emph{always-steer} regime used here the hook is active from the first token, so the correction tracks the board throughout the chain-of-thought; this contrasts with a gated variant that stays dormant until the model closes its \texttt{<think>} block and emits the final-answer \texttt{moves\,=\,[} header, ignoring the speculative move fragments that pepper the reasoning trace. Recovery of accuracy under steering implicates mid-generation degradation as the cause of failure.

For Qwen3.6-27B it does (Table~\ref{tab:steering}). Steering is applied only to the problems the model fails unaided. Even the weakest setting converts $52\%$ of failures into optimal solutions. At the best strength ($\alpha=2$), it lifts the optimal-solve count on the 81 configurations from $33$ ($41\%$) to $59$ ($73\%$). The effect is non-monotonic in $\alpha$, i.e.\ too large a value destabilises generation. The residual failures are rule-violating moves, not parse errors: well-formed but wrong. This is direct causal evidence that the degraded world model limits the model.

For DeepSeek-R1-Distill-Qwen-32B the same intervention is far weaker and non-monotonic (Table~\ref{tab:steering}). At its best ($\alpha=1$) it converts only $6$ of the $72$ failed problems into optimal solutions, against Qwen3.6-27B's $26$ of $48$; one step in either direction it falls to two, and at $\alpha\geq5$ to one. Larger $\alpha$ does not sharpen the plan but corrupts the output: parse errors climb from $29$ of $72$ at $\alpha=0.5$ to $60$ of $72$ at $\alpha=10$, so stronger steering pushes the model toward unparseable traces rather than better ones. The natural explanation is output format: DeepSeek-R1-Distill-Qwen-32B frequently fails to emit a parseable move list at all, a failure a state-restoring nudge to the residual stream cannot reach, and that stronger nudges only exacerbate.

\begin{table}[t]
  \centering
  \setlength{\tabcolsep}{3pt}
  \caption{Activation steering on the problems each model fails unaided (Qwen3.6-27B at layer 28, $N{=}48$; DeepSeek-R1-Distill-Qwen-32B at layer 44, $N{=}72$).}
  \label{tab:steering}
  \small
  \begin{tabular}{llrrrr}
    \toprule
    \textbf{Model} & \textbf{$\alpha$} & \textbf{Opt.} & \textbf{Subopt.} & \textbf{Incor.} & \textbf{Illegal} \\
    \midrule
    \multirow{5}{*}{Qwen3.6-27B} & 0.5 & \textbf{25} & 7  & 0 & 16 \\
    & 1  & \textbf{21} & 11 & 0 & 16 \\
    & 2  & \textbf{26} & 7  & 0 & 15 \\
    & 5  & \textbf{20} & 11 & 0 & 17 \\
    & 10 & \textbf{25} & 3  & 2 & 18 \\
    \midrule
    \multirow{5}{*}{DeepSeek-R1-Distill} & 0.5 & 5  & 0 & 0 & \textbf{67} \\
    & 1  & 6 & 3 & 0 & \textbf{63} \\
    & 2  & 2  & 0 & 1 & \textbf{69} \\
    & 5  & 1  & 0 & 0 & \textbf{71} \\
    & 10 & 1  & 0 & 1 & \textbf{70} \\
    \bottomrule
  \end{tabular}
\end{table}

\section{Discussion}
\label{sec:discussion}

The two streams lead to one narrative: a geometrically structured world model emerges by the end of the prompt in models separated by orders of magnitude in size, degrades over the course of generation, and recovers solution accuracy when restored at inference.

\subsubsection{An emergent, causal world model that is not an artifact of scale.}
The distance-matching probe recovers the joint configuration, and reproduces the Sierpi\'nski geometry of the state graph, at the toy SEP token and at the final prompt token of both frontier models with near-identical fidelity, down to the shared Spearman--Pearson gap that signals inflated separation between the largest-disk sub-triangles (Table~\ref{tab:sep-probe}). That a 6-layer Transformer trained from scratch and a 27B reasoning model encode the state alike argues the world model is a property of the task, not of scale or training regime. We extend the emergent-world-model line~\citep{li2022emergent, toshniwal2022chess, spies2025transformers} from one-move-per-pass game settings into planning solved through extended chain-of-thought. Activation patching closes the loop from correlation to causation: substituting a donor's SEP activation transfers state to the recipient's output in a large majority of pairs (Table~\ref{tab:patching}), so the representation is read, not merely present.

\subsubsection{Degradation through generation reframes the collapse.}
A faithful prompt-time representation does not yield a solution: both models recover the state near-perfectly at Position A yet solve only about half of flat-to-flat instances optimally (Tables~\ref{tab:flatflat} and~\ref{tab:posBC}). Tracking the probe through generation locates the gap: the joint configuration becomes hard to read at the commitment point and a more factored, per-disk encoding partially returns during emission, the large-model counterpart of the toy unified-to-factored shift. This sharpens \citet{shojaee2025illusion}: the collapse is, at least in part, a failure to maintain a representation the model demonstrably had, not the absence of one. Restoring the clean Position-A activation throughout generation nearly doubles Qwen3.6-27B's optimal-solve count (Table~\ref{tab:steering}), direct causal evidence that the degraded world model is the bottleneck. As in Othello-GPT~\citep{li2022emergent}, the steered model acts on the injected configuration, emitting well-formed move lists with essentially no format errors across the swept $\alpha$; its residual failures are illegal \emph{moves}, not malformed output.

\subsubsection{Steering does not transfer.}
For DeepSeek-R1-Distill-Qwen-32B the intervention barely transfers: at its best ($\alpha=1$) only $6$ of $72$ failed problems become optimal, against $26$ of $48$ for Qwen3.6-27B, and the effect collapses at higher strengths (Table~\ref{tab:steering}). The cleaner differentiator is output format: DeepSeek frequently emits no parseable move list at all, and stronger steering makes this worse rather than better. Parse errors rise from $29$ to $60$ of $72$ as $\alpha$ goes from $0.5$ to $10$, so a large share of its failures occur at a stage a state-restoring nudge to the residual stream cannot reach. A representational mismatch, the position-A direction we inject not aligning with DeepSeek's own state code at the steered layer and position, remains possible; separating this from the format-failure explanation is the most important open question we leave to future work.

\subsubsection{Limitations.}
Experiments (other than the baseline) use a single puzzle at a single size ($n{=}4$) Tower of Hanoi's clean recursive structure is what makes its graph a convenient probe target. Whether the story of emergence, then degradation, and finally restoration holds for less geometric state spaces is untested. The two frontier models are both Qwen-derived, so transfer should be read within that caveat. A supervised probe can fit structure the model does not use; we mitigate this with patching and steering, which test causal use directly. Several figures rest on single seeds. The steering method also relies on an external state-tracker that recomputes the true state after each move, feasible here but not obviously scalable to tasks where ground-truth state is expensive or ambiguous. Finally, the technique that restores a correct world model could steer one toward a non-benign target, a capability whose governance deserves attention as such methods mature.

\section{Conclusion}
\label{sec:conclusion}

We asked where a reasoning model's planning failure lives, and found it is neither an absent world model nor a pure decoding fault but a failure to maintain a representation the model demonstrably forms. Our work makes three contributions that, taken together, reframe the reported collapse in reasoning performance.

First, we refocus attention on the flat-to-flat variant of the Tower of Hanoi. Earlier work studied the tower-to-tower task, which by now is saturated for frontier models. By shifting to flat-to-flat, where even the strongest models solve only about half of instances optimally and accuracy drops steeply with disk count, we obtain a hard planning benchmark where world-model maintenance, rather than initial comprehension, is the limiting factor.

Second, we build a direct bridge between small, in-house Transformers and large reasoning models. We train a 6-layer Transformer from scratch on flat-to-flat traces and show that it develops an emergent world model, a linearly decodable Sierpi\'nski embedding that is causally involved in solving the puzzle. We then apply the same probing, patching, and steering toolkit to two frontier models, Qwen3.6-27B and DeepSeek-R1-Distill-Qwen-32B. This parallel design lets us show that the small and large models encode the state with near-identical fidelity at the prompt, degrade in qualitatively similar ways during generation, and differ chiefly in how well they tolerate the degradation.

Third, we locate the source of failure in representation maintenance rather than absence. A linear distance-matching probe recovers the configuration in faithful Sierpi\'nski geometry at the final prompt token of both frontier models, but tracking through generation shows the representation degrades at the commitment point and returns only in factored form during move emission. Restoring it by steering at inference recovers much of the lost accuracy for Qwen3.6-27B, confirming causality, though the intervention barely helps DeepSeek-R1-Distill-Qwen-32B because its failures also involve unparseable outputs that the residual-stream nudge cannot correct. The reported collapse is thus better read as the model forgetting what it knew than as an inability to represent the problem, pointing mitigation toward maintaining state across a long trace rather than toward more inference-time search.

\newpage
\bibliography{aaai2027}

\end{document}